\documentclass{article}

\usepackage[dblblindworkshop,final]{neurips_2026}
\usepackage[utf8]{inputenc}
\usepackage[T1]{fontenc}
\usepackage{hyperref}
\usepackage{url}
\usepackage{booktabs}
\usepackage{amsfonts}
\usepackage{amsmath}
\usepackage{nicefrac}
\usepackage{microtype}
\usepackage[table]{xcolor}
\usepackage{graphicx}

\AtBeginDocument{\setlength{\abovedisplayskip}{5pt}\setlength{\belowdisplayskip}{5pt}}
\newcommand{\xf}{\ensuremath{\rho}}

\title{Fewer Expert Transfers Without Changing Routing: Expert Eviction for
  MoE Offloading}
\workshoptitle{AXIOM 2026: Foundations of Efficient Deep Learning}

\author{%
  Pierre-Luc Bacon\\
  CarbonForge, Mila\\
  Universit\'e de Montr\'eal\\
  \texttt{pierreluc@carbonforge.ai}
  \And
  Diego Calanzone\\
  Mila\\
  Universit\'e de Montr\'eal
  \And
  Jean-Maxime Larouche\\
  CarbonForge\\
  \texttt{jml@carbonforge.ai}
}

\begin{document}

\maketitle

\begin{abstract}
A mixture-of-experts (MoE) language model uses only a few experts per token,
so a server can keep most expert weights in CPU memory and copy an expert to
the GPU when the router requests it. Recent methods train the router to keep
the same experts for several tokens, and treat fewer routing switches as a
proxy for fewer copies. We measure the trade-off directly, in a harness that
performs and times real CPU-to-GPU copies for OLMoE-1B-7B. On an L40S GPU,
every tested policy that holds an expert set fixed and decodes faster than the
least-recently-used (LRU) cache increases next-token loss by at least $0.067$
nats. A change that leaves every routing decision intact does better: when the
cache must evict an expert, evict the one whose next use lies farthest in the
future. This rule copies $2.1\times$ fewer experts than LRU at no loss in
quality and is within $3.5\%$ of the fewest copies possible. It needs future
requests; the next sixteen tokens of demand recover $96\%$ of its reduction,
whereas a rule based on counts of past requests recovers $27\%$.
\end{abstract}

\section{Introduction}
\label{sec:intro}

A mixture-of-experts layer activates only a few of its experts for each
token. For example, OLMoE-1B-7B uses about 1B of its 7B parameters at a time
\citep{muennighoff2024olmoe}. When the expert weights do not all fit on the
GPU, the rest can stay in CPU memory while the GPU holds a smaller cache. When
the router requests an expert that is not cached, the system copies its
weights to the GPU and evicts another expert. The usual rule evicts the least
recently used expert (LRU) \citep{eliseev2023fast,adapmoe2024}, and an open
vLLM pull request implements this design
(\href{https://github.com/vllm-project/vllm/pull/37190}{PR \#37190}). Because
LRU copies whatever the router asks for, the model computes exactly what it
would with every expert on the GPU. The price is that each miss stalls
decoding until its copy completes.

Serving systems shorten these stalls with prefetching, request traces,
CPU/GPU scheduling, or periodic cache updates
\citep{eliseev2023fast,freetoken2026,xue2024moeinfinity,fmoe2025,fiddler2024,kong2024swapmoe}.
They report results on different hardware with different measures of
locality, and models differ in how well they tolerate a restricted expert set
\citep{liang2025local}, so the published numbers do not say which policy gives
the best trade-off between speed and quality.

A second line of work changes the router instead. It keeps the same expert set
active for several tokens and learns when to replace it, which is called
\emph{temporally extended routing}. \citet{shen2026temporally} train an
option-critic controller \citep{bacon2017optioncritic} with a penalty on
switching sets \citep{harb2018waiting}. It lowers the fraction of tokens that
switch sets from over 50\% to under 5\% while retaining up to 90\% of
the base model's accuracy. In
that evaluation, however, every expert stays on the GPU: the router may choose
only within the selected set, but no weights move. Other methods train the
router to reuse experts \citep{remoe2026,melinoe2026}. Both lines treat fewer
switches as a proxy for lower serving cost.

Fewer switches need not mean fewer copies, even with a cache of two slots. The
sequence $A,B,A,B,\ldots$ switches expert at every token but misses only
twice, when $A$ and $B$ first enter the cache. The sequence
$A,A,B,B,C,C,\ldots$ switches half as often, yet each switch requests an
expert that is not cached. A switch is a change in the routing decision,
whereas a copy happens only when the requested weights are absent. When copies
block computation, decoding time is
\begin{equation}
\label{eq:cost}
\text{ms/token} = c + \xf \cdot \tau,
\end{equation}
where $c$ is compute time, $\xf$ is the number of experts copied per token,
and $\tau$ is the time to copy one expert. Holding an expert set fixed lowers
$\xf$ by restricting which experts the router may use, which can change the
output. Changing which expert the cache evicts lowers $\xf$ without changing
any routing decision. The hardware sets $\tau$, which varies fivefold across
the GPUs in Table~\ref{tab:tau}.

\begin{figure}[t]
\centering
\includegraphics[width=\linewidth]{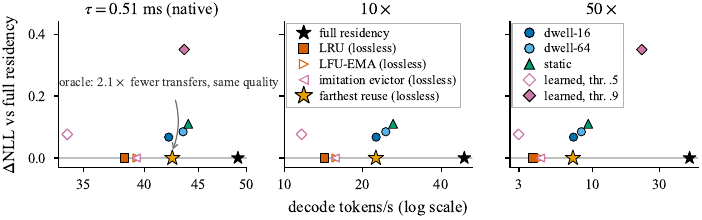}
\caption{\textbf{Quality loss against decoding speed as expert copies become
more expensive.} Left: measured L40S throughput for policies run in the
harness; the farthest-future points combine their counted copies with the
measured compute time. Middle and right: Equation~\eqref{eq:cost} with copies
$10\times$ and $50\times$ slower; among six policies checked by direct
measurement, it picks the measured winner at both costs (\S\ref{sec:tau}).
Points at $\Delta$NLL${}=0$ preserve every routing decision. At the native
cost, no policy that holds an expert set fixed decodes faster than LRU without
at least $0.067$ nats of extra loss, while farthest-future eviction loses
nothing and copies $2.1\times$ fewer experts.}
\label{fig:frontier}
\end{figure}

\looseness=-1
We compare both approaches in a PyTorch harness that performs real, blocking
CPU-to-GPU copies. Half of the frozen model's MoE layers keep 32 of their 64
experts on the GPU and the rest in CPU memory, and quality, copy counts and
decoding time come from the same run. At the native L40S cost, every tested
policy that decodes faster than LRU increases next-token negative
log-likelihood (NLL) by at least $0.067$ nats (Figure~\ref{fig:frontier}). Our
contributions are:
\begin{itemize}\setlength{\itemsep}{1pt}\setlength{\topsep}{2pt}

\item \textbf{Measured offloading.} We compare six cache policies, including
a reimplementation of the learned controller, under real CPU-to-GPU copies.
Equation~\eqref{eq:cost}, fed with measured compute and copy times, picks the
fastest of them when copies are made $10\times$ and $50\times$ slower
(\S\ref{sec:setup}--\ref{sec:tau}).

\item \textbf{A lossless alternative to LRU.} Evicting the expert whose next
use lies farthest in the future copies $2.1\times$ fewer experts than LRU
without changing the model's computation. Sixteen tokens of future demand
recover $96\%$ of this reduction (\S\ref{sec:belady}).

\item \looseness=-1 \textbf{Fewer switches do not give fewer copies.} At four
of six thresholds, a simpler schedule matches or beats the learned
controller, and the other two lower traffic only with large losses in quality.
A router fine-tuned for simulated cache hits copies as many experts as before
(\S\ref{sec:learned}).

\end{itemize}

\vspace{-4pt}
\section{Measurement setup}
\label{sec:setup}
\vspace{-2pt}
We freeze OLMoE-1B-7B \citep{muennighoff2024olmoe} with a router trained to
reuse experts, following \citet{remoe2026}. Eight of its sixteen MoE layers are
offloaded, chosen by restricting one layer at a time
(Appendix~\ref{app:harness}). In each offloaded layer, all 64 bf16 experts
($12.6$\,MB each) stay in pinned CPU memory and the GPU holds $C{=}32$ of them.
Every policy decodes the same sixteen held-out sequences of at most 384 tokens,
teacher-forced with batch size one. We record next-token NLL, tokens per
second and expert copies, leaving out the copies that first fill the cache. A
policy that copies every requested expert produces the same outputs as a run
with all 64 experts on the GPU, and that run is within $5\times10^{-3}$ NLL and
$5\%$ throughput of the model's own MoE block. Code, traces and raw results are
at
\href{https://github.com/carbonforgeAI/moe-expert-eviction}{\texttt{github.com/carbonforgeAI/moe-expert-eviction}}.

The model and router stay frozen, and the policies differ only in how they
manage the 32 GPU slots. LRU copies each missing expert at once and evicts the
least recently used expert that the current token does not need, so it
preserves every routing decision. The other policies change the cache only at
scheduled times. Between changes they copy nothing, and the router must choose
among the cached experts, which can change the output.

\emph{Dwell-$k$} holds the cache fixed for $k$ tokens, then refills it with the
experts requested most often in a recent window: $8k$ tokens for Dwell-16 and
$k$ tokens for Dwell-64 (Appendix~\ref{app:harness}). \emph{Static} fills the
cache from the first 64 tokens and keeps it for the rest of the sequence.
\emph{Learned} is our implementation of \citet{shen2026temporally}. At each
token, a termination head outputs a score $\beta$, and the cache is replaced
when $\beta$ exceeds a threshold, so higher thresholds replace it less often;
a selection head ($\mu$) chooses the new set. Training penalizes switches,
changed experts and misses with hand-set weights rather than the measured
$\tau$, and both heads run inside the timed decoding step. The original trained
heads are not available, so this tests our implementation of the method, not
the published checkpoint.

Copies block computation in our harness, so copy time adds to compute time as
in Equation~\eqref{eq:cost}. A system that overlaps copies with computation
would instead approach $\max(c,\xf\tau)$, which shrinks the benefit of a better
cache on fast links. The measured $\tau$ belongs to this pinned-memory copy
path, not to the GPU in general.

\vspace{-4pt}

\section{Results}

\label{sec:results}

\vspace{-2pt}

Three questions organize the results. Does holding an expert set fixed make
decoding faster at matched quality (\S\ref{sec:tau})? How many copies can a
better eviction rule remove without changing any routing decision
(\S\ref{sec:belady})? Do the learned approaches reduce the copies actually made
(\S\ref{sec:learned})? An aggregate summary from a second text shard, whose raw
logs were not kept, orders the policies the same way
(Appendix~\ref{app:replication}).
\vspace{-3pt}

\subsection{Quality and throughput across expert-transfer costs}

\label{sec:tau}

\begin{table}[t]

\centering

\caption{\textbf{Left:} quality, speed and steady-state copies per token
($\xf$) on the L40S, where one expert copy takes $\tau=0.51$\,ms. The
farthest-future throughput is computed from Equation~\eqref{eq:cost}; the
others are measured. Shaded policies preserve every routing decision and match
the NLL of the all-64-expert baseline. \textbf{Right:} copy time and bandwidth
across GPUs; raw logs were kept only for the L40S.}

\label{tab:frontier}\label{tab:tau}
\scriptsize
\begin{minipage}[t]{0.62\linewidth}
\centering
\begin{tabular}{l c | c c}
\toprule
 & quality & \multicolumn{2}{c}{serving cost} \\
\cmidrule(lr){2-2}\cmidrule(lr){3-4}
Policy & $\Delta$NLL & tok/s & $\xf$ \\
\midrule
\rowcolor{black!8} All 64 experts cached & $+0.000$ & $49.0$ & $0.0$ \\
\rowcolor{black!8} LRU (lossless) & $+0.000$ & $38.3$ & $9.6$ \\
\rowcolor{black!8} \emph{Farthest-future eviction} & $+0.000$ & $42.5$ & $4.6$ \\
Dwell-16 (window $8\times$) & $+0.067$ & $42.2$ & $4.5$ \\
Dwell-64 (window $1\times$) & $+0.085$ & $43.5$ & $3.9$ \\
Static (64-token profile) & $+0.110$ & $44.0$ & $3.4$ \\
Learned, thr.\ $0.5$ & $+0.076$ & $33.8$ & $12.3$ \\
Learned, thr.\ $0.9$ & $+0.350$ & $43.6$ & $0.9$ \\
\bottomrule
\end{tabular}
\end{minipage}\hfill
\begin{minipage}[t]{0.35\linewidth}
\centering
\begin{tabular}{lrr}
\toprule
 & \multicolumn{2}{c}{transfer cost} \\
\cmidrule(lr){2-3}
GPU & $\tau$ (ms) & GiB/s \\
\midrule
H100 / H200 (PCIe~5) & $0.24$ & $49.7$ \\
A100-SXM4 & $0.48$ & $24.5$ \\
L40S & $0.51$ & $23.0$ \\
L4 (PCIe~4$\times$8) & $1.03$ & $11.4$ \\
RTX 8000 (PCIe~3) & $1.17$ & $10.0$ \\
\bottomrule
\end{tabular}
\end{minipage}
\end{table}

On the L40S, LRU matches the NLL of the all-64-expert baseline but decodes at
$38.3$ instead of $49.0$ tokens/s, about $22\%$ slower. Policies that hold the
cache fixed copy fewer experts and can decode faster, but every tested policy
that beats LRU's speed raises next-token NLL by at least $0.067$ nats. At this
PCIe~4 copy cost, holding an expert set fixed buys speed with quality; it does
not improve on LRU for free.

\looseness=-1
Which policy is fastest depends on the platform. The copy count $\xf$ depends
on the policy and the token sequence, whereas $c$ and $\tau$ must be measured
on each machine; given both, Equation~\eqref{eq:cost} predicts the fastest
policy. We test that prediction by repeating every copy $10$ or $50$ times for
six policies, and the predicted winner is the measured winner at both costs.
At $50\times$, the equation predicts that Dwell-16 and Dwell-64 decode
$2.0$--$2.5\times$ faster than LRU, and that threshold $0.9$ is the fastest
policy that restricts routing. Dwell-64 and threshold $0.9$ were not among the
six policies run at these costs, so the runs confirm which policy wins, not
the accuracy of the predicted times in general. A deployment must measure $c$
and $\tau$ itself and decide how much loss in quality it accepts.

\vspace{-3pt}

\subsection{A better eviction rule cuts transfers without changing routing}

\label{sec:belady}

\vspace{-2pt}

When the cache is full and a missing expert must come in, the only remaining
choice is which cached expert to evict. That choice changes future misses but
not the current token's computation. Belady's rule \citep{belady1966study}
evicts the expert whose next use lies farthest in the future. It needs the
future requests, so we apply it offline to the requests recorded in the LRU
run; because eviction never changes routing, these are exactly the requests
the model makes under the new rule. Farthest-future eviction lowers copies from
LRU's $9.6$ per token to $4.6$, a $2.1\times$ reduction, with every routing
decision unchanged.

Belady's proof of optimality assumes one request at a time, whereas each MoE
token requests eight experts at once, so the rule is not guaranteed optimal
here. A lower bound of $14{,}587$ copies on our trace, against the rule's
$15{,}087$, places it within $3.5\%$ of the fewest possible
(Appendix~\ref{app:math}).

A deployable rule must instead predict which cached expert will go unused for
the longest. Knowing only the requests of the next sixteen tokens recovers
$96\%$ of the reduction from LRU to the full-lookahead rule. A rule that ranks
experts by exponentially decayed counts of past requests recovers $27\%$, and a
learned scorer that also sees router probabilities recovers $32$--$33\%$
(Appendix~\ref{app:evict}).

\vspace{-3pt}

\subsection{Learned routing does not improve the measured trade-off}

\label{sec:learned}

\vspace{-2pt}

Across six thresholds, the learned controller replaces its cache every $2.9$
to $73$ tokens on average. At thresholds $0.4$ through $0.7$, a simpler policy
gives better quality at the same traffic or less traffic at the same quality.
At threshold $0.5$, for example, the controller and Dwell-16 raise NLL by
similar amounts ($0.076$ and $0.067$), but the controller copies $2.7\times$ as
many experts, and that comparison leaves out the time spent running its two
heads. Thresholds $0.8$ and $0.9$ lower traffic only by raising NLL $0.238$ and
$0.350$ nats.

The selection head predicts well over short spans. When the controller picks
a new set about every three tokens, NLL rises by only $0.030$ nats; when it
keeps each set for 73 tokens on average, NLL rises by $0.350$, more than the
$0.110$ of a static cache chosen from the first 64 tokens. The selector
therefore seems to predict near-term demand, while the termination head keeps
each prediction too long. Retraining the controller with measured copy time in
its objective does not reach the trade-off of the simple policies in a
one-seed sweep. In an aggregate 5,000-question MMLU summary, all three learned
seeds also score below Dwell-32 at the same switch rate, though the
per-question outputs were not kept (Appendix~\ref{app:controller}). The simple
policies restart at each 384-token sequence and the controller does not, so
longer sequences could narrow the gap.

\looseness=-1
The second learned approach changes the router itself. Following
\citet{carmoe2026}, we fine-tune it with GRPO \citep{shao2024deepseekmath}, a
reinforcement-learning method, rewarding the hit rate of a simulated 16-slot
LRU cache at one layer. The training logs report a higher simulated hit rate,
but in the harness the fine-tuned model copies $32{,}050$ experts, against
$31{,}915$ before fine-tuning. At the trained layer and cache size alone, the
counts are $10{,}963$ and $10{,}954$. Adding a cache-size tag to the prompt
increases traffic, and a supervised fine-tuning control stays close to the base
model (Appendix~\ref{app:workingset}).

\vspace{-4pt}
\section{Limitations and conclusion}
\label{sec:limitations}
\vspace{-2pt}

\looseness=-1
Evicting the expert needed farthest in the future copies $2.1\times$ fewer
experts than LRU, comes within $3.5\%$ of the fewest copies possible, and
changes no routing decision. It needs future requests: sixteen tokens of
lookahead recover $96\%$ of the reduction, but the rules we built from past
requests and router probabilities recover only $27$--$33\%$, and closing that
gap is an open prediction problem. The approaches that change routing did not
help. The learned controller does not improve the trade-off between speed and
quality at thresholds $0.4$ through $0.7$, its two lower-traffic settings raise
NLL by $0.238$--$0.350$ nats, and a router trained for simulated cache hits
copies as many experts as before. All results come from one model, batch-1
teacher forcing and copies that do not overlap computation; batching and
asynchronous copies may shrink the gains. Within this setting, expert eviction
is a better target for learning than routing.

\bibliographystyle{plainnat}
\bibliography{refs}

\appendix

\section{Harness details}
\label{app:harness}

\subsection{Weight layout and copies}
For each restricted layer, the gate, up and
down projections of all $64$ experts are packed into three pinned host
tensors. The GPU pool holds $C$ slots per projection and a map from expert id
to slot. A refresh copies only the experts that are not already resident,
$|R_{\text{new}} \setminus R_{\text{old}}|$. Copies that fill a sequence's
empty pool are counted separately from steady-state transfers for every
policy; a pool counts as filling until it has been full once. Copy time is
measured by device events placed around each group of copies, and it is part
of the step time. Each sequence's first token and a global warm-up are
excluded from steady-state timing. In the $10\times$ and $50\times$ runs, the
summed copy events and the end-to-end step events cover different warm-up
periods, so a compute time derived as their difference can be negative. We
therefore use those runs to rank policies, not to split time into compute and
copies.

\subsection{Eviction}
Evicting an expert moves no data. The host keeps an
ordered map from expert id to GPU slot, in order of last use; evicting an
expert removes its entry, and the incoming expert's copy overwrites the freed
slot. Copies and expert matrix multiplications are queued on the same CUDA
stream, so the multiplications always see the new weights. A different
eviction rule, such as decayed frequency or a learned score, therefore
changes only the function that picks the victim. More implementation notes
are in the repository's README.

\subsection{Copy bandwidth}
On the L40S, the per-expert copy loop reaches
$23.0$\,GiB/s at this expert size. We treat $\tau$ as a property of this
pinned host-to-device copy path, not as a constant of the GPU.

\subsection{Warm start and window length}
Dwell policies refresh every token
until their demand window first fills. Without this warm start, the first
pool is filled by tie-breaking on expert ids and then frozen, which in an
earlier version of the harness inflated dwell's quality penalty by up to
half. The window length matters: at dwell $8$, a window $8\times$ longer than
the dwell lowers $\Delta$NLL from $+0.150$ to $+0.073$ and the transfer rate
from $10.9$ to $4.8$ experts per token, compared with a window equal to the
dwell.

\subsection{Exactness checks}
In every run, the harness checks that LRU,
dwell-$1$ with a one-token window, and the pool path at $C{=}64$ match the
base model's NLL within $5\times10^{-3}$. The base NLL itself matches an
independent batched evaluation of the same protocol within $4\times10^{-4}$.

\subsection{Restricted layers}
The eight offloaded layers are
$\{4,5,6,7,8,11,12,13\}$. A probe that restricted one layer at a time
selected them, and the set was then fixed to match the layers on which the
learned controller had been trained, so all baselines use the same set. The
probe's raw outputs were not kept, and we make no quantitative claim about
how layers differ.

\subsection{Choosing a policy}
Given the measured $(c_i, \xf_i, \Delta L_i)$ of
each policy and a tolerated quality loss $\varepsilon$, the fastest
admissible policy is $\arg\min_{\Delta L_i \le \varepsilon}\; c_i + \xf_i\tau$.
Each policy's cost is linear in $\tau$, so the winner can change only where
two of these lines cross, at $\tau_{ij} = (c_j - c_i)/(\xf_i - \xf_j)$. This
is why one measurement of $c$ and $\tau$ at startup is enough to pick the
policy at any copy cost.

\section{Farthest-future eviction is close to optimal}
\label{app:math}

\subsection{Setup}
Consider one restricted layer with $N$ experts, pool capacity
$C$ and $k$ experts requested per token ($k \le C$; here $N{=}64$, $C{=}32$,
$k{=}8$). Teacher-forced decoding of a fixed text produces a demand stream
$S_1,\dots,S_T$, where $S_t$ is the set of $k$ experts that the router
selects at token $t$. A schedule is a sequence of resident sets
$R_0=\emptyset, R_1,\dots,R_T$ with $|R_t| \le C$. It is \emph{lossless} when
$S_t \subseteq R_t$ for every $t$: every requested expert is resident when the
token is computed, so the output equals the output with all experts resident.
Let $\widetilde B(R)=\sum_t |R_t \setminus R_{t-1}|$ count all insertions.
Steady traffic is $B(R)=\widetilde B(R)-C$, because the first $C$ insertions
fill the empty cache and every recorded stream reaches capacity. Layers have
separate streams and pools, so totals are sums over layers, and the reported
rate divides by the $T_{\rm eval}=3{,}296$ timed tokens. The harness counts
the whole group of copies at the filling boundary as initial ($4{,}263$ LRU
fills rather than the replay's $4{,}096$); both count $35{,}872$ LRU
insertions in total, and Table~\ref{tab:frontier} uses each artifact's own
steady count.

Because the router is never restricted in a lossless schedule, the demand
stream does not depend on the schedule. The minimum steady traffic
\[
B^*(C) \;=\; \min\,\{\,B(R) \;:\; |R_t| \le C,\; S_t \subseteq R_t \text{ for all } t\,\}
\]
is therefore a function of one recorded trace, and any lossless scheme,
causal or clairvoyant, with or without prefetching, can be compared on it.

\textbf{Proposition.} \emph{Let $B_{\mathrm{greedy}}$ be the steady traffic
of the schedule that fetches only on demand and, when the cache overflows,
evicts the resident expert that is not requested at the current token and
whose next request lies farthest in the future (an expert never requested
again counts as farthest). Let $B_{\mathrm{flat}}$ be the optimum of the
classical MIN problem on the stream that lists each token's $k$ experts as
consecutive single requests. Then}
\[
B_{\mathrm{flat}} \;\le\; B^*(C) \;\le\; B_{\mathrm{greedy}}.
\]

\emph{Proof.} The upper bound holds because the greedy schedule is lossless
and feasible. For the lower bound, first note that prefetching never reduces
the number of insertions. Suppose that expert $e$ is inserted at time $s$
although $e\notin S_s$, and let $u>s$ be its next request. If there is no such
$u$, or if $e$ is evicted before $u$, delete that residency interval and its
insertion. Otherwise, remove $e$ from $R_s,\ldots,R_{u-1}$, which delays the
same insertion to $u$; the schedule stays feasible and uses no more capacity.
Each step strictly decreases $\sum_t |R_t|$ and never adds an insertion, so
repeating it ends at a schedule that fetches only on demand. Every recorded
stream requests at least $C$ distinct experts, so subtracting the common
first $C$ insertions preserves the order between schedules. Finally, any
lossless schedule serves the flattened stream at the same cost, so $B^*$ is
at least the MIN optimum of that stream
\citep{belady1966study,mattson1970evaluation}. \hfill$\square$

The lower bound is not tight in general: MIN on the flattened stream may
evict an expert right after its request within a token, which a lossless
schedule may not do. Whether the greedy schedule always attains $B^*$ is
therefore open. Exact dynamic programming over all schedules, prefetching
included, found greedy optimal on every one of $4.2\times10^4$ random
instances (uniform, Zipf-distributed and bursty demand, including the served
ratio $k/C = 0.25$--$0.33$).

\subsection{Bounds on the recorded trace}
Choosing the order of requests
within each token that gives the tightest bound for each stream,
$B_{\mathrm{flat}} = 14{,}587$ and $B_{\mathrm{greedy}} = 15{,}087$ ($4{,}810$
and $4{,}985$ on the held-out sequences). The farthest-future count reported
throughout is thus within $3.5\%$ of the optimum. The two bounds coincide on
$14\%$ of the $128$ (layer, sequence) streams, where greedy is exactly
optimal, and the median gap per stream is $2.9\%$. Because
$B_{\mathrm{greedy}}$ is an upper bound on $B^*$, the $2.1\times$ reduction of
\S\ref{sec:belady} is conservative; the exact ratio lies in $[2.1, 2.2]$.

The proposition bounds the minimum number of copies for every lossless
policy. Prefetching cannot lower that minimum, although a poor prefetcher can
add copies. Overlapping copies with computation changes when the necessary
copies block execution, so latency can move from $c + \xf\tau$ toward
$\max(c,\xf\tau)$ without changing how many copies are needed.

\section{Learned eviction}
\label{app:evict}

\subsection{Hand-built rules}
The farthest-future rule can be computed offline
on the recorded demand trace (eight restricted layers, sixteen sequences of
different lengths; counts sum over layers). Replaying the trace therefore
gives an exact label for every eviction decision. Hawkeye
\citep{jain2016hawkeye} and Parrot \citep{liu2020parrot} learn processor
cache replacement from the same kind of labels; our setting has $64$
candidates per set, a frozen model, and router probabilities that forecast
the next request. Evicting the expert with the lowest exponentially decayed
use count recovers $26.9\%$ of the gap between LRU and the oracle with no
learning at all ($31{,}776$ steady transfers for LRU, $27{,}285$ for decayed
frequency and $15{,}087$ for the oracle). Evicting by predicted time to next
use, estimated from each expert's mean interval between requests, is
$17.5\%$ worse than LRU: demand comes in bursts, so an expert used recently
is likely to be used again soon. An aggregate summary of a closed-loop run in
the harness agrees with the replay: decayed-frequency eviction moves
$27{,}118$ experts against LRU's $31{,}609$ ($27\%$ of the oracle gap,
within $0.6\%$ of the replay) at the same checked NLL. The same summary
reports that evicting the expert with the lowest current router logit moves
$46{,}079$ experts, $46\%$ more than LRU, because it evicts experts that were
just used.

\subsection{Imitation from past use}
A small scoring model (linear,
or a two-layer MLP) scores every resident expert at each eviction, and the
expert with the highest score is evicted. It is trained to reproduce the
oracle's choice, and its inputs summarize only the expert's past use: time
since last use, use counts decayed at several rates, and the spacing of past
uses. Trained on twelve of the sixteen sequences and run closed-loop on the
other four, the MLP recovers $25\%$ of the gap ($9{,}231$ transfers against
$10{,}635$ for LRU and $4{,}985$ for the oracle). The same MLP trained on
shuffled labels is far worse than LRU ($-136\%$), so the gain comes from the
oracle's labels. The linear model puts nearly all of its weight on the
frequency features and evicts rarely used experts; on the same four
sequences the hand-built decayed-frequency rule scores $28.7\%$. From past
use alone, imitation rediscovers frequency-based eviction.

\subsection{Limited foresight}
\begin{figure}[h]
\centering
\includegraphics[width=0.62\linewidth]{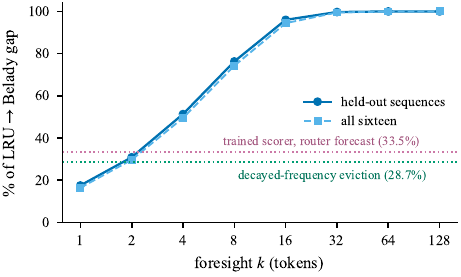}
\caption{Share of the gap between LRU and the oracle that the oracle recovers
when it sees only the next $k$ tokens of requests. The dotted lines are the
causal rules, which correspond to about two tokens of foresight.}
\label{fig:foresight}
\end{figure}

We give the oracle only $k$
tokens of foresight: residents whose next request lies beyond $k$ tokens are
treated as equal, and the least recently used of them is evicted.
Figure~\ref{fig:foresight} shows the result. One token of foresight recovers
$17\%$ of the gap, two tokens $31\%$, four $51\%$, eight $76\%$ and sixteen
$96\%$, and the curve saturates at thirty-two tokens ($99.7\%$; held-out
sequences, with all sixteen sequences within about two points). Nearly all of
the oracle's advantage comes from requests in the next few dozen tokens. The
curve shows how far ahead a causal policy would need to predict; it does not
show that such a prediction is possible from the signals that the server has.

\subsection{Adding router probabilities}
We retrained the scoring model with
router features: each candidate's current probability under the full
softmax, its decayed averages, its trend and its recent maxima. Training used
a separate trace of $48$ sequences ($3.4\times$ as many eviction events), so
all sixteen evaluation sequences are unseen. On the four held-out sequences,
past use alone recovers $26\%$ of the gap, router probabilities alone $25\%$,
and both together $32\%$ with the same two-layer model and $33\%$ with a small
recurrent model (GRU) per candidate. Validation loss on the training trace
selected the two-layer model; we report the GRU but did not select it. Both
beat the decayed-frequency rule ($29\%$) on these four sequences, but on all
sixteen sequences the learned and hand-built rules tie at $25$--$27\%$. The
shuffled-label control again does worse than LRU ($-34\%$). Measured against
the foresight curve, the combined model is worth about two tokens of
foresight, far from the sixteen that would close most of the gap.

\section{The learned controller}
\label{app:controller}

\subsection{Training objectives}
The termination head ($\beta$) and selection
head ($\mu$) are trained on top of the frozen language model with one of two
objectives. The differentiable objective backpropagates next-token NLL
through the soft mixture $(1-\beta)\,\text{keep}+\beta\,\text{refresh}$ and
adds penalties for refreshes, changed experts, expected cache misses and
undecided values of $\beta$, together with a KL term that keeps the routing
close to the original router. The policy-gradient objective applies REINFORCE
to the same decisions, with a reward that combines language-model loss,
refresh rate, rate of changed experts and cache cost, and an entropy bonus.
Both objectives therefore count switches, changed experts and misses, but
with hand-set weights rather than the measured copy time $\tau$.

\subsection{Objective priced by measured copy time}
We retrained the
heads with the traffic penalty set to $e\cdot\tau$, using the measured $\tau$
and otherwise the original configuration, and swept the exchange rate $e$
between quality and time (one seed, same data and number of steps). At $e=0$
the controller refreshes almost every token and moves $73$ experts per token
at full quality. At $e=0.005$ it stays close to that up to threshold $0.8$; at
threshold $0.9$ it lowers traffic from $65.8$ to $41.5$ experts per token for
$+0.071$ nats. At $e=0.02$, the inference threshold becomes a smooth control
from $62$ down to $4.8$ experts per token, and $e=0.08$ nearly reproduces the
curve of the original heads, whose hand-set weight of $0.05$ per changed
expert corresponds to $e\approx0.1$. No price reaches the frontier of the
simple policies: at $4.0$ experts per token the retrained controller loses
$0.16$ nats, while Dwell-64 loses $0.085$ at $3.9$ and farthest-future
eviction loses nothing at $4.6$. Pricing the objective moves the controller
along its own curve but does not close the gap to the frontier.

\subsection{MMLU accuracy}
Table~\ref{tab:mmlu} reports aggregate run summaries
on $5{,}000$ questions, scored by the answer letter, at a matched switch rate
of $3.1\%$ (dwell $32$); base accuracy is $48.4\%$. Per-question outputs and
the trained heads were not kept. At threshold $0.5$ (higher switch rates, $12$
to $16$ transfers per token), the three seeds retain $89.6\%$, $86.5\%$ and
$86.8\%$. Earlier evaluations on $500$ questions spread seed-level retention
over $82$--$96\%$, consistent with their $\pm4.7$-point $95\%$ intervals.

\begin{table}[h]
\centering
\caption{MMLU accuracy at a matched $3.1\%$ switch rate ($n{=}5{,}000$;
aggregate summaries). The $\pm1.3$ is one standard error across three
training seeds.}
\label{tab:mmlu}
\small
\begin{tabular}{lrr}
\toprule
Policy & Accuracy & Retention \\
\midrule
Free routing & $0.484$ & $100.0\%$ \\
Fixed set, oracle selection & $0.481$ & $99.5\%$ \\
Fixed set, chosen from recent use & $0.430$ & $88.9\%$ \\
Learned $\beta\mu$, seed 0 & $0.422$ & $87.3\%$ \\
Learned $\beta\mu$, seed 1 & $0.415$ & $85.7\%$ \\
Learned $\beta\mu$, seed 2 & $0.401$ & $82.8\%$ \\
\midrule
Learned $\beta\mu$, mean of 3 seeds & $0.413$ & $85.2\% \pm 1.3$ \\
\bottomrule
\end{tabular}
\end{table}

\section{A router fine-tuned for cache hits}
\label{app:workingset}

\subsection{Method}
Starting from the public OLMoE-1B-7B-0125-Instruct
checkpoint, the router is fine-tuned with GRPO \citep{shao2024deepseekmath}
following \citet{carmoe2026}. During training, a $16$-slot LRU cache of
experts is simulated at one middle layer over the model's own generations,
and the reward at each token is that cache's hit rate. One variant adds a
prompt tag \texttt{[CACHE\_SIZE=$c$]} so that the cache size can be chosen at
inference time. LoRA adapters on the router, expert and attention
projections are merged into the base weights before measurement. The control
is a model fine-tuned on the same data by supervised learning (SFT) without
the cache reward. An aggregate training summary reports a higher simulated
hit rate at the trained layer; the simulator's raw output was not kept.

\subsection{Protocol}
We use the main protocol of
Appendix~\ref{app:harness}: the same sixteen teacher-forced sequences on the
same GPU. Serving uses lossless LRU, which keeps each model's selected
experts, so steady transfers are the whole comparison. We measure two
operating points: the main one ($8$ restricted layers, $C{=}32$) and the
method's own training setting (the trained layer only, $C{=}16$). The tagged
variant's NLL is not comparable to the others, because the tag tokens are
also scored.

\begin{table}[h]
\centering
\caption{Steady expert transfers (lower is better) for routers fine-tuned
for cache hits, under lossless LRU serving, under lossless serving with
decayed-frequency (LFU) eviction, and at the method's training setting.}
\label{tab:workingset}
\small
\begin{tabular}{lcccc}
\toprule
Objective & Base NLL & LRU & LRU$+$LFU eviction & Trained layer ($C{=}16$) \\
\midrule
Base (untuned)           & $1.382$ & $31{,}915$ & $27{,}275$ & $\mathbf{10{,}954}$ \\
SFT control              & $1.350$ & $\mathbf{31{,}686}$ & $\mathbf{27{,}107}$ & --- \\
Cache-hit reward         & $1.397$ & $32{,}050$ & $27{,}355$ & $10{,}963$ \\
\quad $+$ cache-size tag & ---     & $33{,}101$ & $29{,}063$ & $11{,}223$ \\
\bottomrule
\end{tabular}
\end{table}

\subsection{Results}
The SFT control moves $0.7\%$ fewer experts than the
untuned base, so fine-tuning alone changes little. The reward-trained
variants do not reduce traffic ($+0.4\%$, and $+0.1\%$ at the trained layer
and cache size), and the cache-size tag increases it by $2.5$--$3.7\%$.
Decayed-frequency eviction, by contrast, removes $12$--$15\%$ of the
transfers of every variant without changing the selected experts. Two
differences between the instruments could explain why the simulated gains
do not appear in the measured counts: the reward and its reported gains come
from the same simulator of a single layer, and they were computed on the
model's own generations, while our measurement uses teacher forcing on fixed
text.

\section{Replication and uncertainty}
\label{app:replication}

Aggregate summaries of a second, disjoint text shard (a different offset in
the same corpus, same protocol; base NLL $0.9297$) report the same
orderings. Dwell with a longer window beats dwell with a window equal to the
dwell and the static profile ($+0.075$ to $+0.092$ nats against $+0.115$ to
$+0.142$). The learned controller at threshold $0.5$ matches the heuristics'
quality with about $2.7\times$ as many transfers, threshold $0.9$ again
gives low traffic at a high quality cost, and LRU remains lossless. The raw
logs of this shard were not kept. On the main shard, per-sequence NLLs are
logged for every policy. The points of Figure~\ref{fig:frontier} are
token-weighted aggregate $\Delta$NLL, while the per-sequence spread weights
every sequence equally, so it is not an uncertainty estimate for those
points; we show no error bars and make no claim of statistical significance.
Transfer counts are deterministic given the trace.

\end{document}